\documentclass[10pt,journal,compsoc]{IEEEtran}
\usepackage{amsmath,amssymb,amsfonts}
\usepackage{graphicx}
\usepackage{textcomp}
\usepackage{xcolor}
\usepackage{multirow,algorithm}
\usepackage[algo2e]{algorithm2e}
\usepackage{algorithm,algpseudocode}
\usepackage{subcaption}
\usepackage{paralist}

\ifCLASSOPTIONcompsoc
  \usepackage[nocompress]{cite}
\else
  \usepackage{cite}
\fi

\ifCLASSINFOpdf
\else
\fi
\begin{document}
%
\title{A Metaheuristic-based Machine Learning Approach for Energy Prediction in Mobile App Development}
%
%
%
%

\author{Seyed Jalaleddin Mousavirad
         and Luís A. Alexandre
\IEEEcompsocitemizethanks{\IEEEcompsocthanksitem Seyed Jalaleddin Mousavirad is with the Universidade da Beira Interior, Covilhã, Portugal.\protect\\
\IEEEcompsocthanksitem Luís A. Alexandre is with Universidade da Beira Interior and NOVA LINCS, Covilhã, Portugal.
}
\thanks{Manuscript received ...; revised ...}}

%
%

\markboth{Journal of \LaTeX\ Class Files,~Vol.~14, No.~8, August~2015}%
{Shell \MakeLowercase{\textit{et al.}}: Bare Demo of IEEEtran.cls for Computer Society Journals}
%



\IEEEtitleabstractindextext{%
\begin{abstract}
Energy consumption plays a vital role in mobile App development for developers and end-users, and it is considered one of the most crucial factors for purchasing a smartphone. In addition, in terms of sustainability, it is essential to find methods to reduce the energy consumption of mobile devices since the extensive use of billions of smartphones worldwide significantly impacts the environment. Despite the existence of several energy-efficient programming practices in Android, the leading mobile ecosystem, machine learning-based energy prediction algorithms for mobile App development have yet to be reported. Therefore, this paper proposes a histogram-based gradient boosting classification machine (HGBC), boosted by a metaheuristic approach, for energy prediction in mobile App development. Our metaheuristic approach is responsible for two issues. First, it finds redundant and irrelevant features without any noticeable change in performance. Second, it performs a hyper-parameter tuning for the HGBC algorithm. Since our proposed metaheuristic approach is algorithm-independent, we selected 12 algorithms for the search strategy to find the optimal search algorithm. Our finding shows that a success-history-based parameter adaption for differential evolution with linear population size (L-SHADE) offers the best performance. It can improve performance and decrease the number of features effectively. Our extensive set of experiments clearly shows that our proposed approach can provide significant results for energy consumption prediction.  
\end{abstract}

}

\maketitle

\IEEEdisplaynontitleabstractindextext

%
\IEEEpeerreviewmaketitle

\IEEEraisesectionheading{\section{Introduction}\label{sec:introduction}}

%
%
%
%

\IEEEPARstart{D}{evelopers} are very concerned about how their Apps affect the battery life of phones. One of the most common reasons Apps receive bad reviews in App stores is excessive energy consumption~\cite{smartphone_03,smartphone_04}. Developers are aware of the energy consumption and often ask for help fixing it, even though they rarely get satisfactory advice~\cite{smartphone_05,smartphone_06}. In addition, mobile device manufacturers publish guidelines for developers with the aim of increasing energy efficiency. Also, energy consumption is one of the critical elements affecting mobile device consumers' pleasure. Battery life is identified as the most crucial aspect influencing smartphone purchase decisions in a recent poll of 1898 smartphone users in the US~\cite{smartphone_03,smartphone_04}. Energy consumption by a smartphone has become a concern in recent years. For instance, it has been estimated that 9 out of 10 consumers experience low battery anxiety~\cite{smartphone_05,smartphone_06}. 

From a sustainability perspective, finding ways to make mobile devices use less energy is essential. Also, the billions of phones that are used today have a significant impact on the world. For instance, our use of digital devices, like smartphones and tablets, will likely considerably affect global warming more than the aviation industry~\cite{smartphone_07}.

Optimising, or even just analysing, how much energy mobile devices use is a complicated and time-consuming task for users and/or developers. Keeping track of how much energy an App uses requires many tests in different situations and on several devices, which is a time-consuming task. Developers might use several monitoring tools, often leading to context-specific findings. Android is also a heterogeneous platform. There are likely 3,3 billion Android smartphone users in 190 countries around the world\footnote{http://www.bankmycell.com/blog/how-many-android-users-are-there}. Developer's perception of how hardware works is limited to their own devices and without the right tools and skills, they cannot compare how their Apps use energy or see how Apps work on other devices or in different settings and situations. Also, the energy behaviour of the same App changes depending on how it is used, such as in different OS versions or with varying components of hardware turned on. This is something that has to be taken into account when making a comparison.

In the past few years, several studies~\cite{Energy_APP08, Energy_APP_09,Energy_APP_10,Energy_APP05,Energy_APP06,Energy_APP07} have considered energy-aware programming trends in Android and tried to find better alternatives, while automatic energy consumption prediction has not been seriously discussed. For instance, \cite{smartphone_08} used proxy metrics, such as CPU usage and the number of bytes written to memory, to investigate their suitability as predictors for the
energy consumption of a music streaming application on a mobile device. They examined several experiments on two Apps, Spotify music and podcast App, to find how these proxies were affected by the energy consumption. In another study~\cite{smartphone_09}, a large-scale user study was performed to measure the energy
consumption characteristics of 15500 BlackBerry smartphone
users. By making a large-scale data set, they made the Energy Emulation Toolkit (EET), which lets developers compare the energy needs of their Apps to real user energy traces. In another work, \cite{smartphone_10} used machine learning (ML) and smartphone environment data to determine if a smartphone's next unlock event can be predicted. They showed that it is possible to predict when the next unlock event will happen by doing a 2-week field study with 27 people, leading to improve accuracy and saving energy by using only software-related background data. They suggest reducing energy consumption by using short-term predictions to reduce needless executions or starting computation-intensive tasks, like OS updates, when the phone is locked. For example, by guessing when the next phone open will happen, smartphones can collect sensor data or prepare content in advance to improve the user experience for the next time the phone is used. In one recent study, \cite{JPEG_Postdoc01} indicated two proxies based on image properties for energy consumption: image file size and image quality. In other words, increasing image file size and image quality can elevate energy consumption. They then proposed a multi-objective approach to address this issue. However, they ignored the user's opinion. Therefore, \cite{JPEG_Postdoc02} included the user opinion by changing the encoding representation. Thus, despite several works, the literature review shows no report on predicting energy consumption using machine learning algorithms.

This paper proposes a metaheuristic-based machine learning approach for the prediction of energy consumption in smartphones. Our proposed algorithm benefits from a histogram-based gradient boosting classification machine (HGBC) for the modelling process and a metaheuristic-based approach for the hyper-parameter tuning and feature selection process. HGBC is an ensemble machine learning algorithm that uses the data histogram for a gradient boosting-based model. We selected the HGBC classifier since this method provided the best results in our initial experiments and compared it to 25 classification algorithms. Our proposed metaheuristic-based machine learning algorithm has two different tasks: finding adequate features (feature selection) and hyper-parameter tuning. In the feature selection step, we aim to find suitable features because they can enhance the prediction performance and reduce the number of features that form the state of a smartphone. To this end, we proposed a two-part real-valued encoding representation. The first part is dedicated to the feature selection process, while the second is assigned to the HGBCM hyper-parameters. Since our proposed metaheuristic-based machine learning algorithm is algorithm-independent, we selected 12 well-established and state-of-the-art algorithms for the search strategy. Among them, a success-history-based parameter adaption for differential evolution with a linear population size reduction (L-SHADE) performs best and is selected as the final approach.  


Therefore, the main contributions of this paper are as follows:
\begin{itemize}
	\item We proposed a machine learning-based approach for energy prediction in mobile App development.  This work is the first study for energy prediction in mobile App development. 
	\item Our machine learning algorithm benefits from an ensemble algorithm to boost the results (HGBC algorithm).
	\item We introduced a new metric, energy consumption per minute (ECPM), for assessing the energy consumption of a smartphone.
	\item We proposed a metaheuristic approach for simultaneous feature selection and hyper-parameter tuning.
	\item Since our proposed metaheuristic approach is algorithm-independent, we applied our approach to 12 different search strategies. Among them, L-SHADE is selected as the final algorithm.
	\item We proposed a novel representation for our proposed metaheuristic approach so that it can perform both tasks effectively, feature selection and hyper-parameter tuning, and in a parallel way. 
	\item To the best of our knowledge, the LSHADE algorithm, the best-performing PBMH algorithm compared to others, has yet to be applied for the feature selection problem and parameter-tuning for the HGBC classifier. Therefore, this paper is the first attempt to examine the effectiveness of the LSHADE algorithm for these two problems.  
	\item From a sustainability perspective, this approach can play a vital role in green computing.   
\end{itemize}

The remainder of the paper is organised as follows. Section~\ref{sec:background} briefly describes the background knowledge, while Section~\ref{sec:proposed} explains our proposed algorithm. Section~\ref{sec:exp} provides results and discussion. Finally, Section~\ref{sec:conc} concludes the paper.

\section{Background Knowledge}
\label{sec:background}
This section provides background knowledge regarding histogram-based gradient boosting classification machine, feature selection, and the L-SHADE algorithm.  
\subsection{Histogram-based Gradient Boosting Classification Machine}

\begin{figure*}[tb]
	\centering
	\includegraphics[width=1.5\columnwidth]{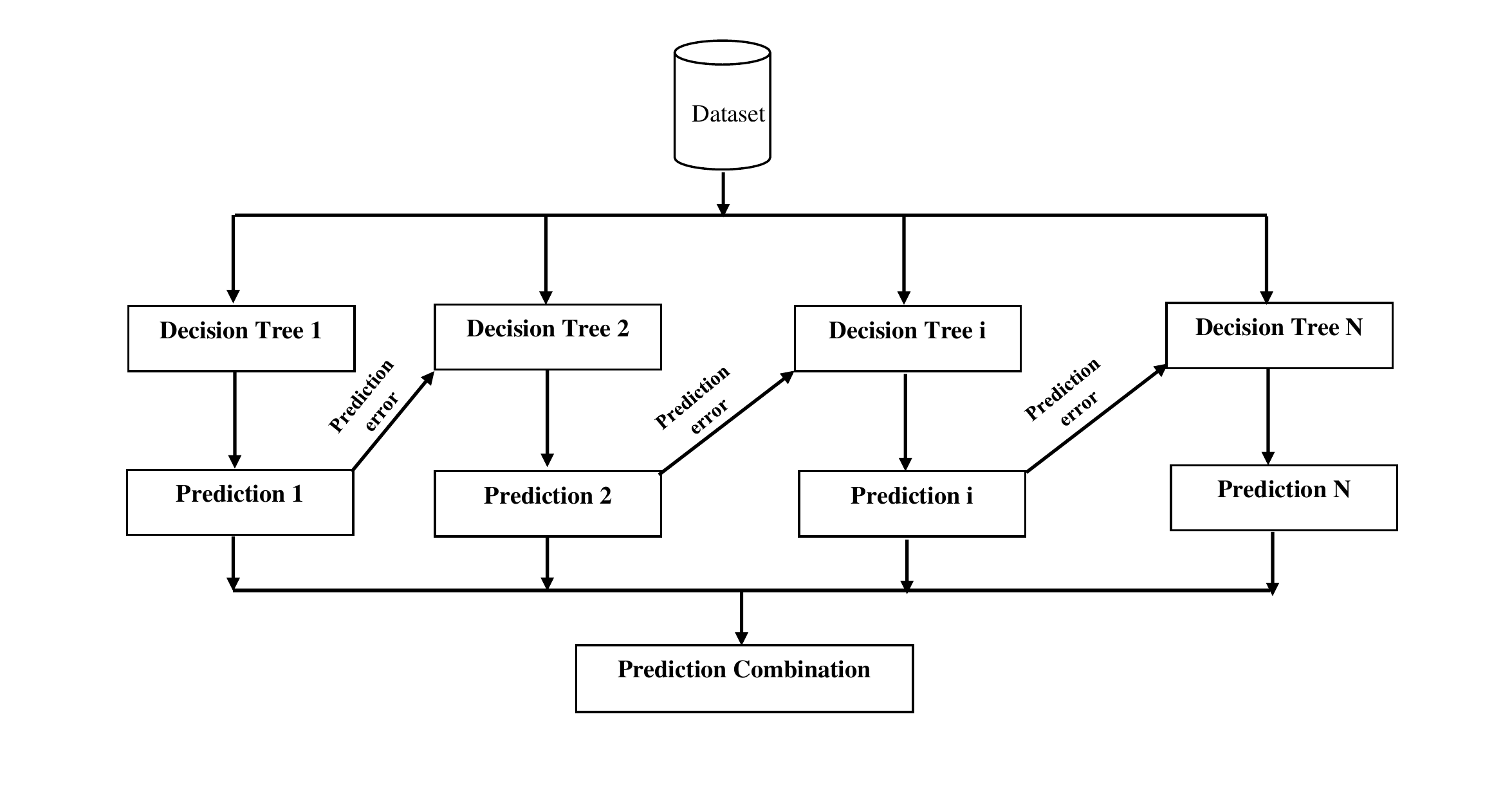}
	\caption{General structure of a gradient boosting ensemble.} 
	\label{fig:gradient}
\end{figure*}

Histogram-based Gradient Boosting Classification Machine (HGBC), Figure~\ref{fig:gradient}, is an approach for building a classification model that combines a gradient boosting machine with a histogram-based algorithm. Gradient boosting is the name given to a group of machine learning models that use simple models, here decision trees, as base learners. A succeeding tree in an
ensemble depends on its ancestors. This is so that the later tree aims at decreasing the misclassified instances performed by the former. During the model training phase, the loss function, which measures the misclassification rate of the entire ensemble, is slowly reduced through this process. After the training phase, a strong committee can be built based on several weak classifiers.

\begin{figure*}[tb]
	\centering
	\includegraphics[width=1.5\columnwidth]{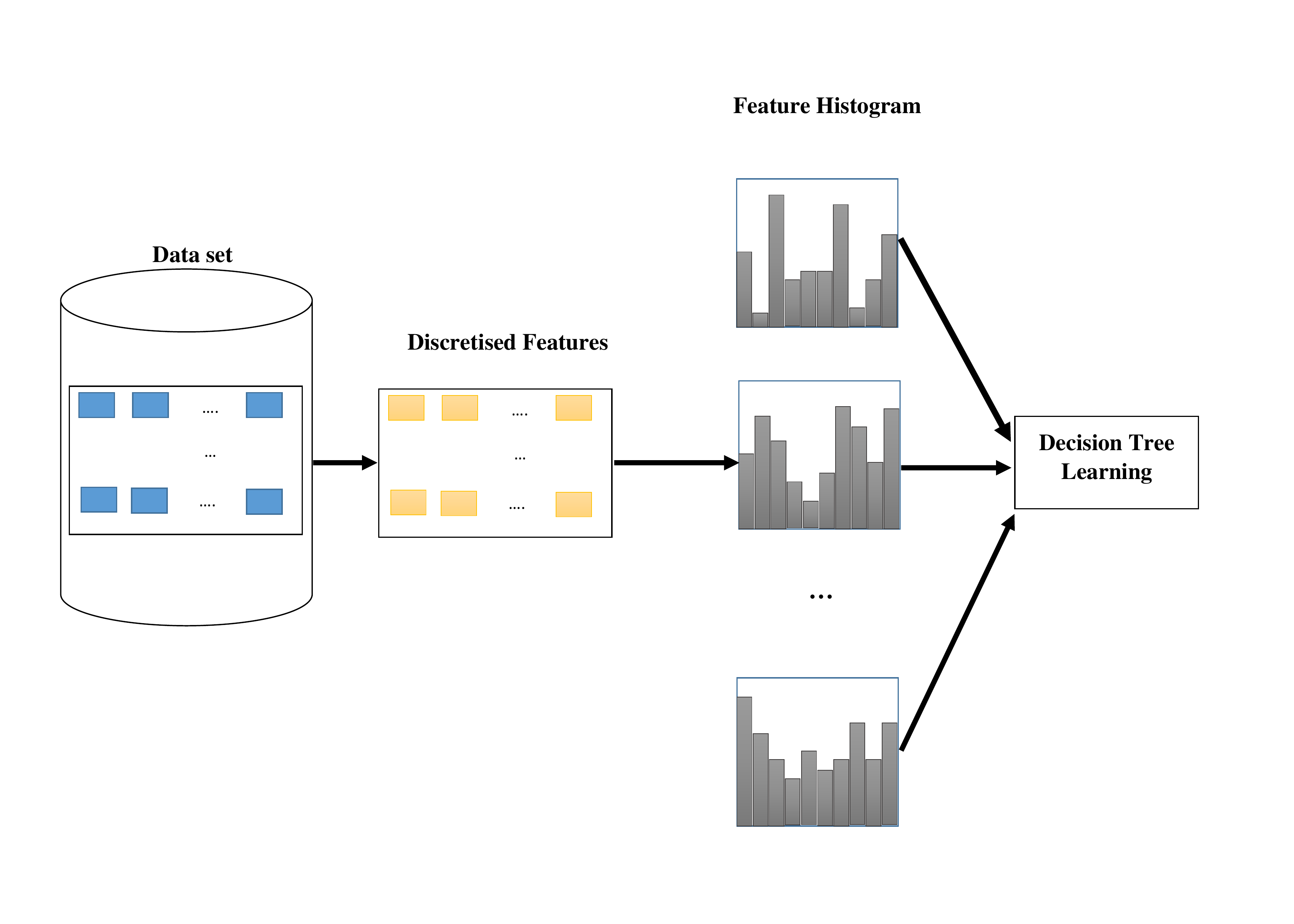}
	\caption{Illustration of the histogram-based algorithm.} 
	\label{fig:hist}
\end{figure*}

In addition, the histogram-based approach, Figure~\ref{fig:hist}, is a method for gradient boosting-based model training, leading to discretising the ranges of continuous features into small bins. These bins are then handled to construct histograms representing the distribution of features' values. The statistical information about the histogram allows for determining the optimal split points for training the base learners. Because the training stage of a decision tree does not necessitate scanning the whole range of features for split point evaluation, the histogram-based technique can significantly lower the computing cost~\cite{HGBCM02}. Additionally, because the training phase is less vulnerable to noise, the histogram-based technique can also improve generalisation~\cite{HGBCM01}.



\subsection{Feature Selection for Classification Tasks}
Feature selection aims to select suitable features for a given data set. Assume that a data set includes $D$ features ($F_{1}, ..., F_{D}$). The feature selection techniques seek to minimise a specified performance function ($H$) by obtaining a feature subset $S$ with $d$ features chosen from the initial feature set ($d \le D$). $H$ typically reflects the classification error rate when selecting features for a classification task.

Mathematically speaking, feature selection aims to find a subset of features $S^{*} \subseteq \mathbb{R}^{D}$ among all possible combinations so that 
\begin{equation}
\label{eq:mutation_shade}
H(S^{*}) \leq H(S_{j}), \forall S_{j} \in \mathbb{R}^{j}
\end{equation}
where $ 1\leq j \leq D$.

Generally speaking, feature selection methods are divided into three leading categories: wrapper, filter, and embedding. Filter methods employ a statistical analysis (e.g. the correlation among features and classes) for selecting the optimal features. In contrast, wrapper methods select a subset of features based on the results of classification algorithms. In addition, embedding methods perform the feature selection task as a part of the learning process. PBMH algorithms such as genetic algorithm (GA)~\cite{FS_GA01}, particle swarm optimisation (PSO)~\cite{FS_PSO01}, and differential evolution (DE)~\cite{FS_DE01} can be used as a wrapper-based feature selection. More details of existing PBMH-based feature selection can be seen in \cite{FS_Survey}.  

\subsection{L-SHADE Algorithm}
\label{Sec:lshade}

In this paper, we applied our strategy to 12 PBMH algorithms. Since the L-SHADE algorithm outperforms other algorithms and is selected as the final algorithm, we explain this algorithm in detail here, while for other algorithms, more details can be seen in the cited publications. 

Differential Evolution (DE), proposed by Storn and Price~\cite{DE_Original}, is one of the most effective PBMH algorithms. The DE algorithm uses three leading operators: mutation, crossover, and selection. Crossover moves the mutant vector and its parent, whereas mutation tries to produce mutant vectors based on the differences among possible solutions. The selection operator makes a greedy choice between the newly produced trial and its parent to pass on the best candidate solution to the subsequent generation. Moreover, the DE method contains three distinct control parameters:  \textit{F}, \textit{CR}, and \textit{NP}. These three parameters represent the scaling factor in the mutation operator, crossover rate, and population size, respectively. The problem-dependent parameter values significantly influence the DE algorithm's performance. One of the state-of-the-art variants of the DE algorithm, success-history-based parameter adaptation for DE (SHADE) ~\cite{SHADE01}, has demonstrated excellent performance over DE. The SHADE algorithm changes the parameters \textit{F} and \textit{CR} adaptively, while the \textit{NP} is fixed. SHADE with Linear Population Size Reduction (L-SHADE)~\cite{LSHADE01} algorithm, the winner of CEC2017 Evolutionary Computation Challenges, adds an adaptive population size to the SHADE algorithm. The main components of the L-SHADE algorithm are described below. 
\subsubsection{External Archive}
In order to maintain diversity, SHADE uses an external archive ($A$). To this end, parent vectors inferior to the trial vector, $u_{i,G}$, are preserved. Then, the combination of the external archive and the current population ($ A \cup P$) is employed in the mutation strategy for updating the position.
\subsubsection{Mutation Strategy}
The SHADE algorithm uses the \textit{current-to-pbest/1} strategy method to update each candidate solution as
\begin{equation}
\label{eq:mutation_shade}
v_{i,G}=x_{i,G}+F_{i}(x_{pbest,G}-x_{i,G})+F_{i}(x_{r1,G}-x_{r2,G})
\end{equation}
where $x_{pbest,G}$ is a randomly selected candidate solution from the top $N \times p$ candidate solutions, and $p \in [0,1]$, where $p$ specifies the greediness of the strategy. In addition, $F_{i}$ denotes the $F$ parameter value by $x_{i}$. $x_{r1,G}$ is a random candidate solution selected from the population, $P$, while $x_{r2,G}$ is a candidate solution selected randomly by $ A \cup P$. The size of the Archive is matched to the population size. 

Moreover, the SHADE method uses an adaptive technique to adjust the parameter $p$ such that each candidate solution has a unique $p$ that is set as
\begin{equation}
p_{i} =rand[\frac{2}{N},0.2].
\end{equation}

\subsubsection{History Based Parameter Adaptation}
A historical memory containing \textit{H} entries helps the SHADE algorithm maintain control over the parameters \textit{CR} and \textit{F}. First, the entire contents of memory are set to 0.5($M_{CR}$ and $M_{F}$). In each iteration, a random integer number, indicating a random entry, $r_{i}$, should be chosen in the range of [1,\textit{H}]. Then, the parameters $CR_{i}$ and $F_{i}$ are defined as
\begin{equation}
CR_{i}=randn_{i}(M_{CR,r_{i},0.1})
\end{equation}
\begin{equation}
F_{i}=randc_{i}(M_{F,r_{i},0.1})
\end{equation}
where $randn_{i} (\mu,\sigma^{2})$ and $randc_{i}(\mu,\sigma^{2})$ are responsible for generating random numbers, with mean $\mu$ and variance $\sigma^{2}$ using the normal and Cauchy distributions, respectively. \textit{CR} and \textit{F} produced values should be constrained to [0,1].

The $CR_{i}$ and $F_{i}$ values used by good candidate solutions are kept in $S_{CR}$ and $S_{F}$. In other words, $S_{CR}$ and $S_{F}$ included the $CR_{i}$ and $F_{i}$ values that successfully generate a trial vector superior to the parent individual. The memory contents should be updated at the end of each iteration as 
\begin{equation}
M_{CR,K,G+1}=\begin{cases}
mean_{WA} (S_{CR}) & if S_{CR} \neq 0 \\
M_{CR,K,G} & \text{otherwise}.
\end{cases} 
\end{equation}

\begin{equation}
M_{F,K,G+1}=\begin{cases}
mean_{WL} (S_{F}) & if S_{F} \neq 0 \\
M_{F,K,G} & \text{otherwise}.
\end{cases} 
\end{equation}
where $K$ shows an index, initialisation by 1, indicating the memory location between 1 and $H$. Furthermore, $mean_{WA}$ and $mean_{WL}$ denote the weighted arithmetic and weighted
Lehmer means.
\subsubsection{LPSR Technique }
The population size, \textit{P}, is not altered by the SHADE method, but the parameters \textit{F} and \textit{CR} are. Smaller populations tend to converge more quickly, while larger populations tend to explore more and slow the convergence rate. In order to reduce the population size throughout the optimisation phase, L-SHADE uses a Linear Population Size Reduction (LPSR) technique, defined by the number of function evaluations. To put it another way, the population size is continuously decreased using LPSR using a linear function as
\begin{equation}
P_{G+1}=round\left[\frac{P^{min}-P^{int}}{Max_{NFE}}.NFE+P^{int}\right]
\end{equation}  
where $P^{min}$ is the minimal population size, which here is 4 because \textit{current-to-pbest/1} strategy calls for a minimum of 4 candidate solutions, $P_{int}$ shows the initial  population size in the first iteration, $Max_{NFE}$ is the maximum number of function evaluations, and $NFE$ is the current number of function evaluations. Also, $P_{G+1}$ denotes the population size in the next generation.

\section{The proposed algorithm}
\label{sec:proposed}
This paper proposes a novel metaheuristic-based machine learning algorithm for energy prediction in mobile App development. To this end, our proposed algorithm employs an HBGC classifier for the modelling process and proposes a metaheuristic algorithm for selecting the relevant features as well as hyperparameter tuning. Our ML-based energy prediction approach can be deployed for android App development in different techniques:  
\begin{itemize}
	\item Clung-App technique: the ML component is part of an App, which causes this component to exist in a smartphone as many Apps on the same smartphone, which is redundant.
	\item Clung-Android technique: in this technique, the ML component is a part of Android as a built-in component. While this technique might be the best, it would require major backers, such as Google, to decide that such a built-in component is a viable option, which is currently not feasible. 
	\item Independent-App technique: the ML component is presented as an independent App. It is challenging to persuade a user to install an additional App on top of the App required by the user.
	\item Web service technique: the ML component acts as a web service in this technique. Therefore, whenever an application intends to use this component, it can send a request to the web service and receive a recommendation. This method works well until major Android backers decide that a built-in ML-based component for energy consumption is necessary.
\end{itemize}

\subsection{Data set}
\label{sec:dataset}
We have used the GreenHub data set~\cite{greenhub_dataset}. This data set includes more than 23 million instances, from over 900 brands and 5,000 models, across 160 countries. It consists of three types of information, including sample data set, device data set, and App processes data set. The sample data has several details on different smartphone settings, while the device data set includes several device features such as brands. In addition, the App processes data set has features related to the installed Apps in each smartphone. 

In this paper, we only used the sample data set and ignored App processes and device data sets. We disregarded App processes because we aim to propose a predictive algorithm independent of the installed applications. In addition, since each geographical region has its native applications, it will make the predictive process challenging. For example, while the popular messenger software in China is $Wechat$, it is $WhatsApp$ in Portugal. Therefore, if these features are used, the data set will be sparse. Also, ignoring this data set will make the results independent of the geographical area. The same circumstances also are applicable to the device data. Therefore, we can say that the results are only dependent on the Android settings.

The sample data set has seven types of features, from which we selected 32 features. The features are listed below, while more details can be seen in \cite{greenhub_dataset}:
\begin{itemize}
	\item Battery details: charger (unplugged or plugged), health, voltage, and temperature,
	\item CPU states: usage, uptime, and sleep time,
	\item Network details: network type, mobile network type, mobile data status, mobile data activity, roaming enabled, wifi status, wifi signal strength, wifi link speed, 
	\item Samples: battery state, battery level, memory free, memory user, network status, screen brightness, screen on,
	\item Settings: Bluetooth enabled, location enabled, power saver enabled, flashlight enabled, NFC enabled,  developer mode
	\item Storage Details: free, total, memory active, and memory inactive.
\end{itemize}

\subsection{Preprocessing}
This step is responsible for transforming or encoding the existing data set so that the machine learning algorithms can easily work on it. The steps are listed below.
\subsubsection{Instance Elimination}
 In this step, we eliminated the instances including “battery-state=charging” since our goal is to predict the energy consumption in the discharging state. Also, the instances with missing values are removed from the data set.

\subsubsection{ECPM Metric }
In this paper, we defined a new metric to show the amount of energy consumed per minute by a smartphone. Energy Consumption Per Minute (ECPM) is defined as
\begin{equation}
ECPM=\frac{BT_{state1}-BT_{state2}}{TS_{state1}-TS_{state2}} \time 60
\end{equation}
where $BT_{state1}$ is the battery level in state 1, while $TS_{state1}$ means the time stamp in state 1. In addition, State 1 and State 2 are two consecutive states. State of a smartphone here means the current settings of a smartphone, such as Bluetooth and Wifi states (on/off). Also, we have two assumptions. First, the current state of a smartphone is not dependent on the earlier states, and second, the smartphone settings have not changed between the consecutive states. ECPM can be used as a metric for energy consumption. Lower values of ECPM show a lower energy consumption. Since we assumed that there are no smartphone settings changes in two consecutive states, if “battery-state=charging” state is between two “battery-state=discharging” states, the corresponding instance is not considered for our new metric. 


\subsubsection{Histogram Analysis}
Since this research considers the energy prediction as a classification problem, here we have assigned each instance to a class based on ECPM metric. The number of selected classes is three, namely safe, warning, and critical status. Assigning each sample into a class depends on the mobile App developer, for example, for one developer, ECPM=0.5 may be high, but for another developer, this value is in a safe state. Since we have no knowledge of developer interest, here, we used a histogram analysis to select classes, although this could easily change later. 
From the histogram, instances with $ECPM < 0.5$ are assigned to the first class, while instances with $ECPM > 1.5$ are set to another class. Also, the remaining instances belong to the third class. 

 
\subsection{Metaheuristic-based HGBC Algorithm}
This paper proposes a metaheuristic-based HGBC algorithm for the modelling process. For this, three issues to be determined are the structure of each candidate solution (representation), the objective function, and the search strategy, which we define in the following.
\subsubsection{Representation of Solutions}
\label{Sec:Rep}
Our proposed PBMH aims to find the optimal features, numbers, and parameters for the HGBC classification algorithm. Therefore, we proposed a two-part candidate solution: a real-valued vector of length $N_{D}+N_{P}$. The first $N_{D}$ elements are in the interval [0,1] and are employed for selecting the features. The remaining elements indicate the $N_{P}$ parameters of the HGBC classification algorithm. 


The first part of our proposed representation is dedicated to the feature selection process, which is a string with a length equal to the number of features (here, it is 32). The vector's cells are initialised at random with real values between 0 and 1. Despite the use of this representation in continuous search space~\cite{FS_PSO02}, we used this representation since the operators in the employed search algorithms are designed for the continuous search space, and the goal of this paper is not to define new operators for discrete search spaces. Therefore, a feature is used when a cell's value is above 0.5. Thus, the feature set can be defined by the following notation: The value of the corresponding cell changes to 1 when a feature is selected and to 0 otherwise. As a result, the string is transformed into a binary vector, where 0 denotes the feature's rejection, and 1 denotes its selection.
  
The second part of our representation is assigned to the HGBC parameters, learning rate ($lr$), minimum leaf nodes ($MnLN$), maximum leaf nodes ($MxLN$), L2 regularisation, and maximum bins ($mb$). The boundary of the search space in these parameters is different from the first part of the representation. The $lr$ parameter is set as a number between 0.001 (as a negligible number close to 0) and 1, while L2 regularisation is a number between 0 and 3. Other boundaries for $MnLN$, $MxLN$, and $mb$ are set as [1,29], [30,100], and [2,255]. Since $MnLN$, $MxLN$, and $mb$ should be integer numbers, after each continuous operation in the search strategy, they are rounded to the nearest integer.   

Therefore, the total length of our used representation is (32+5=37). In addition, this paper aims to find both the optimal number of features and the parameters of the HGBC classification algorithm simultaneously using only one representation strategy.

It is worth mentioning that the search space size is large in our proposed representation. The solution space of the first part of our problem is $2^{32}$. In addition, assume points are located equally with
a distance of 0.01 in $lr$ and L2 regularisation parameters. Therefore, the solution space size in this problem is at least $2^{32} \times 100 \times 300 \times 29 \times 71 \times 254$ which is a large number, and makes this is a challenging process for the optimisation algorithms. 
\subsection{Objective Function}
 The objective function here is responsible for hyper-parameter tuning, and feature selection, simultaneously, which is calculated in three following steps: 
\begin{enumerate}
	\item In the first step to compute the objective function, the features corresponding to the representation should be selected. In other words, the input to the training phase will be only the features that have a value of one in the corresponding representation.
	\item  HGBC classification algorithms should be trained based on the selected features and the hyper-parameters available in the second part of the representation.
	\item The objective function here is calculated based on a classification error, which is defined as
	
	\begin{equation}
	\text{Classification error}=\frac{100}{P} \times \sum_{p=1}^{P} \xi(x_{p})
	\label{Eq_Obj}
	\end{equation}
	with
	\begin{equation}
	\xi(x_{p})=
	\begin{cases}
	1 &  \text{if} \hspace{0.1cm} o_{p}\neq d_{p}\\
	0& \hspace{-0.04cm} \text{otherwise}
	\end{cases}\hspace{2cm}\\
	\label{eq7}
	\end{equation}
	where $P$ is the number of instances, and for each input vector in training data, the corresponding desired output is $d_{p}$, and $o_{p}$ is the predicted output.  
\end{enumerate}

This objective function can simultaneously measure both the feature quality and the adequacy of the hyper-parameters. To have robust results, the objective function is calculated using a $k$-fold cross-validation. 
\subsection{Search Strategy}
This paper aims to find suitable features and hyper-parameters for the HGBC classification algorithm for energy prediction in mobile App development. Since the problem is novel, there is no research on this issue. Therefore, we selected 12 metaheuristic algorithms, including both base and state-of-the-art variants, as follows:
\begin{compactitem}
	\item Random search (RS),
	\item Genetic algorithm (GA)~\cite{GA_original01},
	\item Artificial bee colony (ABC)~\cite{ABC_Main_Paper},
	\item Covariance matrix adaptation evolution strategy (CMA-ES)~\cite{CMA_main_paper},
	\item Evolution strategy (ES)~\cite{ES_main_paper},
	\item Differential evolution (DE)~\cite{DE_Original},
	\item Hierarchical PSO Time-Varying Acceleration (HPSO-TVAC)~\cite{HPSO-TVAC},
	\item Phasor Particle Swarm Optimization (P-PSO)~\cite{Phasor-PSO},
	\item Adaptive Differential Evolution With Optional External Archive (JADE)~\cite{JADE_01},
	\item Success-History Adaptation Differential Evolution (SHADE)~\cite{SHADE01},
	\item Linear Population Size Reduction Success-History Adaptation Differential Evolution (LSHADE)~\cite{LSHADE01}.
\end{compactitem}	

Our experiments (available in Section~\ref{sec:exp}) showed that LSHADE outperforms other algorithms. Therefore, we focus on the LSHADE search strategy and propose LSHADE-HGBC as our final prediction approach, while the details of other strategies can be seen in the cited publications.

LSHADE-HGBC aims to find the optimal features and hyper-parameters of HGBC classification algorithms using the LSHADE algorithm. To this end, like other population-based algorithms, it starts with a set of randomly candidate solutions, and its structure is explained in Section~\ref{Sec:Rep}. Then, the quality of each generated solution is evaluated by Eq.~\ref{Eq_Obj}. The LSHADE algorithm is responsible for finding the optimal solutions based on the operators introduced in Section~\ref{Sec:lshade}. The final solution is an array including both the features selected by the LSHADE-HGBC algorithm and the optimal hyper-parameters for HGBC classifier. These parameters are used for the final evaluations. The proposed algorithm in the form of Pseudo-code is described in Algorithm~\ref{Alg1:proposed}.

\begin{algorithm2e}[tb]
	\caption{LSHADE-FS-HGBCM algorithm in the form of Pseudo-code.}
	\scriptsize
	\SetAlgoLined
	\SetKwInOut{Input}{Input}\SetKwInOut{Output}{Output}
	\Input{ $D$: dimensionality of problem; $NFE_{\max}$: maximum number of function evaluations; $NP_{\max}$: maximum population size; $NP_{\min}$: minimum population size; $H$: Memory size}
	\Output{ $x^*$: the best individual }
	\BlankLine
	$M_{F}=0.5$, $M_{CR}=0.5$, $NFE=0$, $t=1$, $NP=NP_{max}$\\
	Randomly generate initial population $Pop$\\
	Evaluate objective function value of each individual using Eq.~\ref{Eq_Obj} and the new data set generated by the selected features and the hyper-parameters embedded in an individual\\
	$x^{*} \leftarrow$: the best candidate solution from the current population\;
	\While{$NFE<MAX_{NFE}$}{
		$S_{F}=0,S_{R}=0$ \\	
		\For{$i\leftarrow 1$ \KwTo $NP$} {
			Generate random index $r_{i}=rand(1,H)$\\
			Generate $CR_{i}^{t}$ as $randn_{i}(M_{CR,r_{i}},0.1)$ \\
			Generate $F_{i}^{t}$ as $randc_{i}(M_{F,{r_{i}}},0.1)$\\
			Generate $p_{i}^{t}$ as $rand[p_{min},0.2]$\\
			Select $x_{r1}$ randomly from current population \\
			Select $x_{r2}$ randomly from combination of current population and archive \\
			Generate trial vector as $v_{i}^{t}=x_{i}^{t}+F_{i}^{t}(x_{best}-x_{i}^{t})+F_{i}^{t}(x_{r1}-x_{r2})$ \\
			\For{$j\leftarrow 1$ \KwTo $D$} {
				\eIf{$rand_{j}[0,1]<C_{R}$ or $j==j_{rand}$} {
					$u_{i,j}^{t}={v}_{i,j}^{t}$
				}{
					$u_{i,j}^{t}=x_{i,j}^{t}$}
			}
		}
		\For {$i\leftarrow 1$ \KwTo $NP$} {
			\eIf{$f(u_{i}^{t}) \leq f(x_{i}^{t})$} {
				$ x_{i}^{t+1} \leftarrow u_{i}^{t} $ \\
			} {
				$ x_{i}^{t+1} \leftarrow x_{i}^{t} $ \\
			}
			\If{$f(u_{i}^{t}) < f(x_{i}^{t})$} {
				$A \leftarrow x_{i}^{t}$\;
				$S_{CR} \leftarrow CR_{i}^{t}$ \\
				$S_{F} 	\leftarrow F_{i}^{t}$ \\	
			}
		}
		When size of archive exceeds $|A|$, randomly delete individuals so that $|A| \leq |P|$ \\
		\If{$S_{CR}\neq 0$ and $S_{F}\neq 0$} {
			Compute $M_{F,K}^{t+1}=mean_{WL}(S_{F})$ \\
			Compute $M_{CR,K}^{t+1}=mean_{WA}(S_{CR})$ \\
		}
		Perform LPSR as $NP=round[\frac{NP_{max}-NP_{min}}{MAX_{NFE}}.NFE+NP_{max}]$ \\	
		t=t+1 \\
		$x^{best} \leftarrow$: the best candidate solution from the current population\\
			\If{$f(x^{*}) > f(x^{best})$} {
			$x^{*} \leftarrow x^{best}$\\	
		}
		
	}
$N\_Data\leftarrow$ select the optimal features based on $x^{*}$ and create a new data set \\

$Opt\_param \leftarrow$ select the optimal parameters for HGBCM classification algorithms  \\

Apply HGBCM on the $N\_Data$ data set by using the optimal parameters ($Opt\_param$)\\
	\label{Alg1:proposed}
\end{algorithm2e}

\subsection{Strength and Limitations}
This paper examined a wide range of classification algorithms and PBMH algorithms to find the best predictive combination. One of the most important properties of the proposed algorithm is that it can predict effectively without any knowledge regarding the applications installed in a smartphone. This is important because it is much more challenging to ask a user for permission to know all the Apps installed on the smartphone than to obtain an App's settings. Also, the number of Apps for Android is huge, and more Apps are provided daily. Providing a predictive algorithm by ignoring installed software can solve this problem to a large extent. Also, the proposed prediction algorithm is independent of the device. In other words, it can be applied on all brands and models.

Despite the strength of the proposed method, this research needs to address one crucial limitation: permission. Some of these features require permission from a user, so a user may not grant all these permissions to an App. In other words, the value of some features might not be present in the feature set. This case can not be handled by the proposed approach.  
\section{Experimental results and discussion}
\label{sec:exp}
In this section, we provided an extensive set of experiments to assess our proposed algorithm. To this end, we selected two main criteria, accuracy and F-measure.  For the experiments, we used $k$-fold cross-validation (kCV) So that a data set is divided into $k$ almost equal parts. Each time, one part is used as test and the rest as training. This is repeated $k$ times and the statistical results including mean and standard deviation (std.) are reported. The details of the data set we used are available in Section~\ref{sec:dataset}. 

\subsection{Performance Analysis}

In this paper, we proposed 12 PBMH-based HBGC algorithms. All representation and objective functions are selected the same, and only their search strategy differs. The population size and the number of function evaluations, as the stopping criterion, for all algorithms are selected 50 and 800, respectively, while we have used default values for the other parameters given in Table~\ref{tab:parameter}.

The results are given in Table~\ref{tab:meta}. From the table, LSHADE-HGBC provides the best results, while DE-HGBC and RS-HGBC perform worst. Most algorithms, including RS, ABC, CMA, ES, GA, PSO, HPSP, and DE, can not achieve better results than the HGBC algorithm with the default parameters, indicating the poor ability of these search strategies to find the optimal solutions. Among all algorithms, only JADE, SHADE, and L-SHADE, all the improved variants of DE, can enhance the results. In particular, LSHADE-HGBC can improve the classification error by more than 5\%.

Due to the nondeterministic nature of metaheuristic algorithms, non-parametric statistical methods are required to understand the significance of the results. In this case, the alternative hypothesis $H_{1}$ indicates a statistically significant difference among the algorithms, whereas the null hypothesis $H_{0}$ asserts no statistical difference among the algorithms. The $H_{0}$ is the initial statistical assertion, and if the $H_{0}$ is shown to be false, the $H_{1}$ is accepted. To statistically compare the results, we applied the Wilcoxon signed rank test~\cite{tutorial_statistical} at a significance level of 5\% based on the 10CV classification results. The Wilcoxon signed-rank test was chosen since it does not presume normal distributions and is, therefore, safer than the t-test. In addition, the Wilcoxon test is less sensitive to outliers than the t-test~\cite{tutorial_statistical}. Table~\ref{tab:wil_acc} shows the Wilcoxon signed rank test results based on 10CV classification accuracy. From the table, we can observe that LSHADE outperforms others significantly since, in all cases, LSHADE is the winning algorithm, while JADE placed in the second rank because it wins against 10 other algorithms. The third rank belongs to SHADE, which was the winner in 9 cases. The overall subsequent best-performing
algorithms are GA (7 wins, 1 tie, and 3 fails) and ABC (5 wins, 2 ties, and 3 fails). The worst algorithms are DE (11 fails), and RS (1 win and 10 fails). In addition, the same table is obtained for the results of the Wilcoxon signed rank test based on 10CV F-measure. Therefore, due to the limitation page, we did not include this table in the paper.
\begin{table}[tb]
	\centering
	\caption{Parameter settings for the experiments.}
	\label{tab:parameter}
	\begin{tabular}{l|p{2.5cm}|p{3cm}}
		Algorithms                     & Parameter          & Value                                      \\ \hline
		RS                             & -                  & -                                          \\ \hline
		\multirow{3}{*}{GA}            & PC                 & 0.95                                       \\
		& PM                 & 0.05                                       \\
		& selection          & Tournament                                 \\ \hline
		\multirow{4}{*}{ABC}           & limit              & $n_{e} \ \times$ dimensionality of problem \\
		& $n_{o}$          & 50\% of the colony                         \\
		& $n_{e}$          & 50\% of the colony                         \\
		& $n_{s}$          & 1                                          \\ \hline
		ES                             & $\lambda$          & 0.75                                       \\ \hline
		CMA-ES                         & -                  & -                                          \\ \hline
		\multirow{3}{*}{`DE}           & weighting   factor & 0.8                                        \\
		& crossover   rate   & 0.9                                        \\
		& Strategy           & DE/current-to-rand/1/bin                   \\ \hline
		\multirow{4}{*}{PSO}           & C1                 & 2.05                                       \\
		& C2                 & 2.05                                       \\
		& $w_{min}$          & 0.4                                        \\
		& $w_{max}$          & 0.9                                        \\ \hline
		\multirow{6}{*}{HPSO-TVAC}     & C1                 & 2.05                                       \\
		& C2                 & 2.05                                       \\
		& $w_{min}$          & 0.4                                        \\
		& $w_{max}$          & 0.9                                        \\
		& ci                 & 0.5                                        \\
		& cf                 & 0                                          \\ \hline
		\multirow{4}{*}{P-PSO}         & C1                 & 2.05                                       \\
		& C2                 & 2.05                                       \\
		& $w_{min}$          & 0.4                                        \\
		& $w_{max}$          & 0.9                                        \\ \hline
		\multirow{6}{*}{JADE}          & weighting   factor & 0.8                                        \\
		& crossover   rate   & 0.9                                        \\
		& Strategy           & DE/current-to-rand/1/bin                   \\
		& $cr$                 & 0.5                                        \\
		& $P_{t}$                 & 0.1                                        \\
		& $A_{p}$                 & 0.1                                        \\ \hline
		\multirow{4}{*}{SHADE, LSHADE} & weighting   factor & 0.8                                        \\
		& crossover   rate   & 0.9                                        \\
		& Strategy           & DE/current-to-rand/1/bin                   \\
		& Memory   size      & 50  \\ \hline                                      
	\end{tabular}
\end{table}
\begin{table}[tb]
	\centering
	\caption{10CV classification results for different PBMH-based THGBC algorithms. }
	\label{tab:meta}
	\begin{tabular}{l|cccc}
		\hline
		\multirow{2}{*}{Algorithms} & \multicolumn{2}{l}{Accuracy} & \multicolumn{2}{l}{F-measure} \\ \cline{2-5} 
		& Mean          & Std.         & Mean           & Std.         \\ \hline
		HGBC                        & 79.33         & 1.86         & 79.13          & 1.88         \\
		RS-HGBC              & 77.95         & 1.69         & 77.76          & 1.74         \\
		ABC-HGBC             & 79.23         & 1.54         & 79.03          & 1.60         \\
		CMA-HGBC             & 78.68         & 1.69         & 78.46          & 1.70         \\
		ES-HGBC              & 79.13         & 1.43         & 78.88          & 1.54         \\
		GA-HGBC              & 79.49         & 1.81         & 79.30          & 1.87         \\
		PSO-HGBC             & 79.23         & 1.47         & 79.00          & 1.49         \\
		HPSO-HGBC            & 78.78         & 1.76         & 78.60          & 1.81         \\
		PPSO-HGBC            & 78.45         & 0.84         & 78.23          & 0.91         \\
		DE-HGBC              & 77.35         & 1.70         & 77.06          & 1.73         \\
		JADE-HGBC            & 80.18         & 1.40         & 80.08          & 1.38         \\
		SHADE-HGBC           & 80.08         & 1.53         & 80.00          & 1.76         \\
		LSHADE-HGBC         & 80.43         & 1.61         & 80.23          & 1.65         \\ \hline
	\end{tabular}
\end{table}

\begin{table*}[tb]
	\centering
	\caption{Results of Wilcoxon signed rank test based on the 10CV accuracy results. +, -, and = denote that the algorithm in the corresponding row is statistically superior to, inferior to, or equivalent to the algorithm in the corresponding column, respectively. The final column summarises the cumulative wins (w), ties (t), and losses (l) of each algorithm.}
	\label{tab:wil_acc}
	\begin{tabular}{l|cccccccccccc|l}
		& RS & ABC & CMA & ES & GA & PSO & HPSO& PPSO & DE & JADE & SHADE & LSHAD & w/t/l  \\ \hline \hline
		RS     &         & -          & -          & -         & -         & -          & -           & -           & +         & -           & -            & -             & 1/0/10 \\ 
		ABC    & +         &          & +          & =         & -         & =          & +           & +           & +         & -           & -            & -             & 5/3/3  \\
		CMA    & +         & -          &          & -         & -         & -          & =           & =           & +         & -           & -            & -             & 2/2/7  \\
		ES     & +         & =          & +          &         & -         & =          & =           & +           & +         & -           & -            & -             & 4/3/4  \\
		GA    & +         & +          & +          & +         &         & =          & +           & +           & +         & -           & -            & -             & 7/1/3  \\
		PSO    & +         & =          & +          & =         & =         &          & +           & +           & +         & -           & -            & -             & 5/3/3  \\
		HPSO   & +         & -          & =          & =         & -         & -          &           & =           & +         & -           & -            & -             & 2/3/6  \\
		PPSO   & +         & -          & =          & -         & -         & -          & =           &           & +         & -           & -            & -             & 2/2/7  \\
		DE     & -         & -          & -          & -         & -         & -          & -           & -           &         & -           & -            & -             & 0/0/11 \\
		JADE  & +         & +          & +          & +         & +         & +          & +           & +           & +         &           & +            & -             & 10/0/1 \\
		SHADE  & +         & +          & +          & +         & +         & +          & +           & +           & +         & -           &            & -             & 9/0/2  \\
		LSHADE& +         & +          & +          & +         & +         & +          & +           & +           & +         & +           & +            &             & 11/0/0 \\ \hline
	\end{tabular}
\end{table*}

\subsection{Comparison of Different Variants}
\label{sec:variant}
This paper proposes a PBMH-based algorithm for energy prediction in an Android smartphone. The PBMH algorithm is responsible for finding the hyper-parameters of HGBC and feature selection. In this section, we aim to compare different variants of the proposed algorithm. In the first variant (FS-HGBC), the PBMH algorithm is responsible for only the feature selection process. In contrast, our proposed algorithm only can tune the hyper-parameters without any feature selection step in the second variant (THGBC). Also, we selected the LSHADE algorithm since it performs best. For FS-HGBC and THGBC, we changed the representation of solutions. In other words, FS-HGBC only has 32 elements, while THGBC has 5 parameters, which has caused the size of the search space to become much smaller in both variants.

The results can be seen in Table~\ref{tab:variants}. By comparison between FS-HGBC and HGBC, we can observe that the results are almost the same, but it is imperative to mention two points: firstly, the number of features has decreased, which means that with fewer features, the algorithm was able to achieve the same performance, and secondly, the stability of the results, based on the standard deviation, has also improved (from 1.86 to 1.49 for accuracy and from 1.88 to 1.50 for the F-measure). The number of selected features after the feature selection process is 24, and some features, such as wifi signal speed and roaming enabled, are eliminated. Comparison between HGBC and THGBC indicates that the hyper-parameter tuning can enhance the classification error by more than 5\%, showing the positive effect of hyper-parameter tuning. In addition, the results of LSHADE-HGBC are better than THGBCM and FS-HGBC, meaning that the feature selection process can improve the final performance.    
\begin{table}[tb]
	\centering
	\caption{10CV classification results for different variants of our proposed algorithm.}
	\label{tab:variants}
	\begin{tabular}{l|cccc}
		\hline
		\multirow{2}{*}{Algorithms} & \multicolumn{2}{l}{Accuracy} & \multicolumn{2}{l}{F-measure} \\ \cline{2-5} 
		& Mean          & Std.         & Mean           & Std.         \\ \hline
		HGBC                      & 79.33         & 1.86         & 79.13          & 1.88         \\
		FS-HGBC                    & 79.43         & 1.49         & 79.22          & 1.50         \\
		THGBCM                      & 80.43         & 1.61         & 80.23          & 1.65         \\
		LSHADE-HGBC                   & 81.02         & 1.34         & 80.76          & 1.42         \\ \hline
	\end{tabular}
\end{table}

\subsection{Comparison of Different Classifiers}

In this experiment, we selected several classification algorithms for the modelling process. This experiment aims to find the best classification algorithms and justifies choosing the HGBC classifier as the base classifier of our proposed algorithm. The evaluated classifiers are linear discriminant analysis (LDA), quadratic discriminant analysis (QDA), logistic regression (LR), passive aggressive classifier (PAC), stochastic gradient descent (SGD), decision tree (DT), Naive Bayes (NB), $k$-nearest neighbour (KNN), multi-layer perceptron (MLP), support vector classification (SVC), and five ensemble classifiers including adaboost, random forest (RF), extra tree classifier (ETC), gradient boosting classifier (GBC), and histogram gradient boosting classifier (HGBC). Also, we used DT with two different splitting criteria, accuracy (DT-AC) and gini-index (DT-GI), three different NBs, including Gaussian NB, Bernoulli NB, and multinomial NB, four different $k$ for the KNN algorithm ($K=3, 5, 7, 9$), two different structures for the MLP (with 100 and 200 neurons in the hidden layer), four different kernels for the SVC, including, linear, polynomial, RBF, and sigmoid, and two splitting criteria for the RF algorithm, accuracy (RF-AC) and gini-index (RF-GI). Therefore, we tested 26 different classifiers for this experiment.

The 10CV results can be seen in Table~\ref{tab:classifier}. From the table, we can observe that LDA, QDA, linear models, NBs, and SVCs perform worst. In particular, the worst classifiers are SVC with polynomial kernel and PAC classifier. Ensemble classifiers provide the best results. Among the ensemble classifiers, HGBC works best, while adaboost performs worst.  It is worthwhile to mention that minimum accuracy/ F-measure for HGBC is 77.13/77.00, that is higher than maximum accuracy/F-measure for other algorithms. As a result, according to the merits of HGBC, we selected HGBCM as the final classifier for our proposal.

\begin{table}[tb]
	\centering
	\caption{Comparison of different classifiers.}
	\label{tab:classifier}
	\footnotesize
	\begin{tabular}{lcccc}
		\hline
		Algorithms                                   & Accuracy &      & F-measure &      \\ \cline{2-5} 
		& Mean     & Std. & Mean      & Std. \\ \hline
		 Discriminant Analysis &          &      &           &      \\ \hline
		LDA                                          & 56.83    & 1.71 & 56.25     & 1.79 \\
		QDA                                          & 50.24    & 1.87 & 46.63     & 2.46 \\ \hline
		Linear Models                                &          &      &           &      \\ \hline
		LR                                           & 56.71    & 1.89 & 55.74     & 1.94 \\
		PAC                                          & 48.23    & 7.47 & 45.46     & 7.41 \\
		SGD                                          & 55.31    & 2.20 & 53.60     & 2.78 \\ \hline
		Decision Trees                               &          &      &           &      \\ \hline
		DT-AC                                        & 67.13    & 1.40 & 66.97     & 1.40 \\
		DT-GI                                        & 67.06    & 1.72 & 66.88     & 1.74 \\ \hline
		Naive Bayes                                  &          &      &           &      \\ \hline
		Gaussian NB                                  & 45.04    & 1.03 & 37.37     & 1.32 \\
		Bernoulli NB                                 & 52.79    & 1.23 & 52.12     & 1.35 \\
		Multinomial NB                               & 50.60    & 1.78 & 48.87     & 1.66 \\ \hline
		`K- Nearest Neighbors                        &          &      &           &      \\ \hline
		KNN (K=3)                                    & 62.41    & 2.10 & 62.26     & 2.06 \\
		KNN (K=5)                                    & 61.59    & 1.88 & 61.35     & 1.88 \\
		KNN (K=7)                                    & 60.94    & 2.18 & 60.43     & 2.22 \\
		KNN (K=9)                                    & 60.30    & 1.90 & 59.72     & 1.89 \\ \hline
		Multi-layer neural network                   &          &      &           &      \\ \hline
		MLP-V1                                       & 62.25    & 1.82 & 61.70     & 2.00 \\
		MLP-V2                                       & 65.26    & 1.57 & 64.98     & 1.76 \\ \hline
		Support Vector Machines                      &          &      &           &      \\ \hline
		SVC (linear Kernel)                          & 56.23    & 1.84 & 54.51     & 1.94 \\
		SVC (Polynomial Kernel)                      & 47.48    & 0.62 & 36.55     & 1.26 \\
		SVC (RBF Kernel)                             & 55.38    & 1.50 & 52.82     & 1.56 \\
		SVC (Sigmoid Kernel)                         & 54.01    & 1.56 & 50.76     & 1.75 \\ \hline
		Ensemble Classifiers                         &          &      &           &      \\ \hline
		Adaboost                                     & 63.61    & 2.36 & 63.52     & 2.38 \\
		RF-AC                                        & 77.16    & 1.61 & 76.82     & 1.70 \\
		RF-GI                                        & 77.24    & 1.90 & 76.83     & 2.00 \\
		ETC                                          & 62.61    & 2.46 & 62.43     & 2.38 \\
		GBC                                          & 70.49    & 1.79 & 70.22     & 1.83 \\
		HGBC                                       & 79.33    & 1.86 & 79.13     & 1.88 \\ \hline
	\end{tabular}
\end{table}

\subsection{Comparison of LSHADE-based Feature Selection Algorithm with the Conventional Algorithms}

\begin{figure}[tb]
	\centering
	\includegraphics[width=.9\columnwidth]{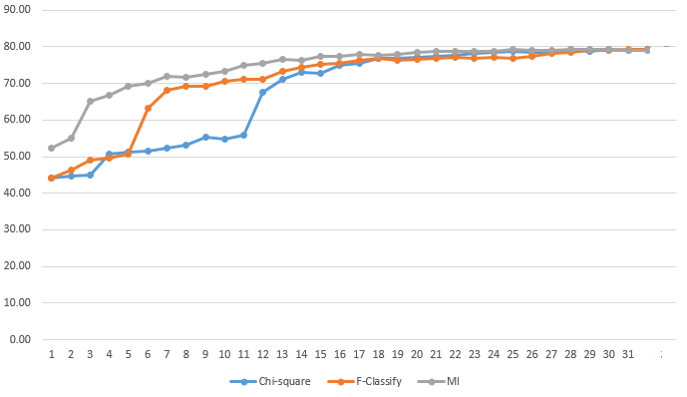}
	\caption{The effect of conventional feature selection algorithms with different number of features in terms of accuracy.} 
	\label{fig:accuracy}
\end{figure}

\begin{figure}[tb]
	\centering
	\includegraphics[width=.9\columnwidth]{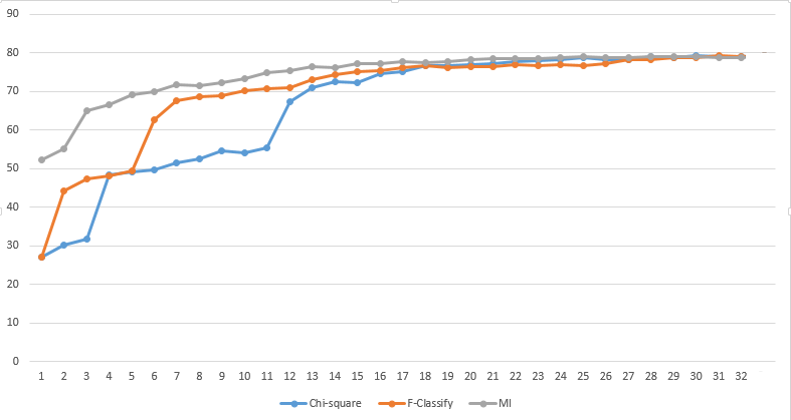}
	\caption{The effect of conventional feature selection algorithms with different number of features in terms of F-measure.} 
	\label{fig:fmeasure}
\end{figure}

In section~\ref{sec:variant}, we showed that the proposed LSHADE-based feature selection could provide the same results with the HGBC classifier but with less features. This section compares this with several conventional feature selection algorithms, including chi-square (CS)~\cite{chi_FeatureSelection}, F-classify (FC)~\cite{Anova_FeatureSelection}, and mutual information (MI)~\cite{MI_FeatureSelection}. Since one of the input parameters of the conventional algorithms is the number of features, we run the experiments with all features. Figures~\ref{fig:accuracy} and \ref{fig:fmeasure} show the accuracy and F-measure results for different selected features. From the Figures, the highest accuracy and F-measure are obtained when the number of features is between 28 and 32 and the performance is almost the same. Therefore, we selected 28 features for comparison. The results can be seen in Table~\ref{tab:conv}. However, the results of the proposed feature selection algorithm are the same as the conventional ones; it can find the number of features automatically. In addition, the proposed algorithm can achieve these results with fewer features. Therefore, we can say that the proposed feature selection performs better than the conventional feature selection algorithms.

\begin{table}[tb]
	\centering
	\caption{Comparison of LSHADE-based feature selection with several conventional algorithms.}
	\label{tab:conv}
	\begin{tabular}{l|ccc}
		\hline
		Algorithms  & \multicolumn{1}{l}{Accuracy} & \multicolumn{1}{l}{F-measure} & \multicolumn{1}{l}{Number of selected features} \\ \hline
		CS     & 78.66                        & 78.66   & 28                      \\
		FC      & 78.95                        &    78.76     & 28                 \\
		MI & 79.33                        & 79.13    & 28                     \\ 
		FS-HGBC        & 79.33                        & 79.13 & 24 \\ \hline
	\end{tabular}
\end{table}

\section{Conclusion}
\label{sec:conc}
Energy consumption is one of the crucial factors for both developers and users of mobile phone Apps, so the quality of an App is associated with the amount of energy it consumes. Knowing information about the future state of a smartphone's energy consumption can enable the developer to manage energy consumption. Therefore, this paper proposes a metaheuristic-based machine learning for energy prediction in mobile App development. After some preprocessing of the data set, we introduced a new metric called Energy Consumption Per Minute (ECPM) for evaluating the energy consumption in an Android device. Then, we proposed a metaheuristic-based histogram-based gradient boosting classification machine (HGBC) for the modelling process. In this paper, the metaheuristic algorithm tackles two different issues: selecting the proper features (feature selection) and tuning the parameters of the HGBC classifier. Our extensive experiments clearly show that the proposed approach can provide satisfactory results. Also, we suggested deploying this approach in real-world problems in four ways: clung-App, clung-android, independent-app, and web service although currently the latter technique is the best approach. Regarding sustainability, our approach can play a crucial role in green computing since the energy consumption of digital devices, like smartphones, will likely considerably affect global warming further than the aviation industry~\cite{smartphone_07}. 

Despite the satisfactory performance of the proposed approach, this work can be extended in the future by addressing the following issues:
\begin{itemize}
	\item The proposed approach employs the combination of the HGBC classification algorithm and LSHADE metaheuristic algorithm for the prediction purpose. However, other algorithms, such as deep learning-based approaches, might have the potential to provide better results.
	\item This paper assumes that all the values in the used data set are present, while some features might be unavailable in practice. Considering this issue in the future is another direction.
	\item This paper employs an ensemble algorithm for the prediction, while the ensemble of solutions found by LSHADE is another possible direction to improve the results.
\end{itemize}

\ifCLASSOPTIONcompsoc
  \section*{Acknowledgments}
\else
  \section*{Acknowledgment}
\fi

This work was financed by FEDER (Fundo Europeu de Desenvolvimento Regional), from the European Union through CENTRO 2020 (Programa Operacional Regional do Centro), under project CENTRO-01-0247-FEDER-047256 – GreenStamp: Mobile Energy Efficiency Services.

This work was supported by NOVA LINCS (UIDB/04516/2020) with the financial support of FCT-Fundação para a Ciência e a Tecnologia, through national funds.

\ifCLASSOPTIONcaptionsoff
  \newpage
\fi

\bibliography{jalal}
\bibliographystyle{plain}
\end{document}